\documentclass{article}
\usepackage{mathtools}
\usepackage{url}
\usepackage{xcolor}
\usepackage{hyperref}
\usepackage{multirow}
\usepackage{graphicx}
\usepackage{booktabs} 
\usepackage{amsmath,graphicx}

\usepackage[preprint]{spconf} 
\copyrightnotice{\copyright IEEE 2022}
\toappear{Published in ICASSP 2022 - {\it scheduled for 22-27 May 2022 in Singapore.}}

\newcommand{\argmax}[1]{\underset{#1}{\operatorname{arg}\,\operatorname{max}}\;}

\def\mt{{\sc MT}}
\def\ad{{\sc AD}}
\def\tts{{\sc TTS}}
\def\pa{{\sc PA}}

\title{ISOMETRIC MT: NEURAL MACHINE TRANSLATION FOR AUTOMATIC DUBBING}

\name{Author(s) Name(s)\thanks{Thanks to XYZ agency for funding.}}
\address{Author Affiliation(s)}
\name{Surafel M. Lakew, Yogesh Virkar, Prashant Mathur, Marcello Federico}
\address{Amazon}

\address{Amazon AI \\
{\small \tt \{surafelm|yvvirkar|pramathu|marcfede\}@amazon.com}
}

\begin{document}
\ninept
\maketitle
\begin{abstract}
Automatic dubbing (AD) is among the machine translation (MT) use cases where translations should match a given length to allow for synchronicity between source and target speech. For neural MT, generating translations of length close to the source length (e.g. within $\pm10\%$ in character count), while preserving quality is a challenging task. 
Controlling MT output length comes at a cost to translation quality, which is usually mitigated with a two step approach of generating N-best hypotheses and then re-ranking based on length and quality. This work introduces a self-learning approach that allows a transformer  model to directly learn to generate outputs that closely match the source length, in short {\em Isometric MT}. In particular, our approach does not require to generate multiple hypotheses nor any auxiliary ranking function. We report results on four language pairs (\textit{English $\rightarrow$ French, Italian, German, Spanish}) with a publicly available benchmark. Automatic and manual evaluations show that our method for Isometric MT outperforms more complex approaches proposed in the literature.
\end{abstract}

\section{Introduction}
\label{sec:intro}
Reaching the global audience is a primary factor for audio-visual content localization. Automating the task of localization requires translation of source language speech and a seamless integration of the target language speech with the original visual content~\cite{federico_speech--speech_2020}. Recent developments in  AD~\cite{oktem2019,federico_speech--speech_2020,federico_evaluating_2020} have focused on achieving {\it isochrony}, a form of synchronization at the level of speech utterances~\cite{chaume:2004}. The AD architecture proposed in~\cite{federico_speech--speech_2020}, includes an \mt~model that translates source transcriptions to a target language, followed by a prosodic alignment (\pa)~module which segments the translation into phrases and pauses according to the prosody pattern in the source speech. Finally, a text-to-speech (\tts) module synthesizes the target language speech for a final rendering with the original audio background and visual content.

For a \tts~module, to generate a natural sounding speech in synchrony with the source utterances, it is important that the length of the translated script should match that of the source.\footnote{Following~\cite{federico_speech--speech_2020}, length of input text in characters is directly proportional to the duration of the \tts~utterance.} 
With a goal of generating translations that match the source length,~\cite{saboo_integration_2019,Lakew19} have proposed approaches that bias the output length of MT. In a subsequent work~\cite{lakew2021verbosity}, the authors proposed a better approach, 
where the main idea is to generate translations that fall within a $\pm10\%$ range of the source length in character count. They also confirmed that generating translations within this range easily allow the \pa~and the \tts~modules to adjust the target speech to a natural sounding speaking rate. \cite{lakew2021verbosity} is state of the art approach for \textit{MT with length control}, but it depends on a two step process {\it i}) $N$-best hypotheses (where $N$=50) generation and {\it ii}) a re-ranking step which interpolates the model scores and length ratios (translation to source) to find the best hypothesis in the $N$-best space. In this work, we propose a self-learning based approach whose improvements are orthogonal to the re-ranker's improvements and when combined we achieve state of the art results on a public benchmark.

In statistical \mt, self-learning~\cite{UeffingHS07} has been investigated to augment training data with pseudo bi-text. 
In neural \mt, the most commonly used self-learning approach is back-translation~\cite{sennrich2015improvingMono}, that leverages a reverse ({\it target $\rightarrow$ source}) model to generate pseudo bi-text from a target language monolingual data, to train the desired {\it source $\rightarrow$ target} direction. Subsequent works~\cite{edunov2018understanding,he2019revisiting,caswell2019tagged},
have shown variants of self-learning using a backward and forward translation settings.

This work proposes a self-learning approach that applies a controlled generation of pseudo bi-text with a length constraint, to model isometric \mt. 
To assess the effectiveness of the proposed approach, we perform evaluation on a speech translation data~\cite{MUSTC:2019}, in four language directions exhibiting different degree of target to source length ratio: {\it English $\rightarrow$ French, Italian, German, Spanish}. Specifically, our contributions are:
\begin{itemize}
  \item We propose a self-learning based approaches to learn \mt~model that can generate isometric translations.
  \item We compare our proposed approach and show it is on par or better than previous state of the art for controlling \mt~output length, without using multiple hypothesis generation and re-ranking function.
  \item We introduce a new \mt~evaluation mechanism leveraging TTS (\textit{i.e, generating audio from translations for rating by subjects}), and metrics to measure the acceptability rate of isometric translations, particularly for an \ad~use case. 
\end{itemize}

The rest of this work, discusses \mt~and output length control approaches $\S$\ref{sec:concise_mt_ad}, followed by a description of our proposed isometric \mt~$\S$\ref{sec:concise_mt_selflearning}, experiment and evaluation settings $\S$\ref{sec:experiments}, and finally results and discussion of our findings $\S$\ref{sec:results}.

\section{Background}
\label{sec:concise_mt_ad}

\begin{figure*}[t!]
    \centering
    \includegraphics[scale=0.99,trim={1.0cm 25.9cm 3cm 1.2cm},clip]{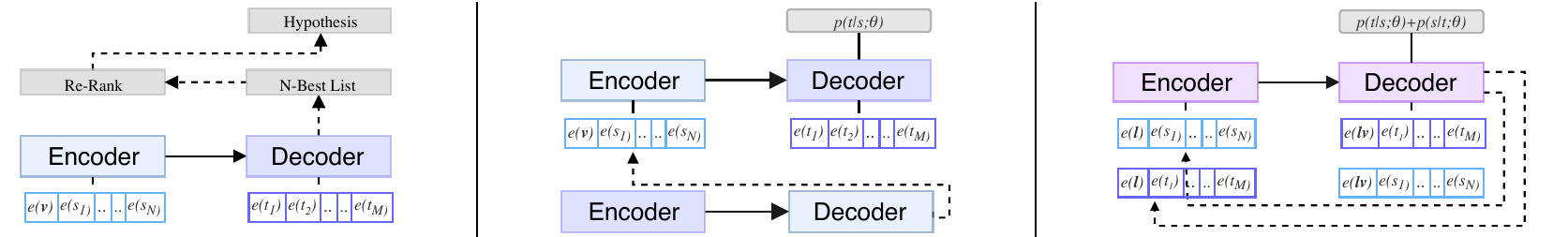}
    \caption{\mt~output length control approaches; (\textit{left}) current SOTA using length token+$N$-best hypotheses re-ranking~\cite{lakew2021verbosity}, (\textit{center}) our proposed offline self-learning using a reverse \mt~(\textit{bottom}) trained in isolation to improve the desired (\textit{top}) forward \mt~direction, and (\textit{right}) the online self-learning variant using a single bi-directional \mt. \textit{Broken} arrows show test time hypotheses generation (\textit{left}), and self-learning training data generation (\textit{center, right}). For simplicity, we show embedding lookup function $e()$ for input tokens, where $e(v)$ for length token, $e(l)$ for language token, and $e(vl)$ for combination of length and language token.}
    \label{fig:concise_mt}
\end{figure*}

\subsection{Neural Machine Translation}
For a language pair with parallel data $\mathcal{D}=\{(s_i, t_i)$: $ i=1,\ldots,N\}$, 
an \mt~model parameterized with $\theta$, trains to maximize likelihood on the training sample pairs as, 

\begin{equation}
    L(\theta) = \argmax{\theta} \sum_{i=1}^N log\ p(t_i|s_i,\theta)
    \label{eq:mt_max_likelihood}
\end{equation}

\noindent
The model considers the source context ($s$) and previously generated target language tokens ($t_{<k}$), when forming the prediction ($t$),

\begin{equation}
    p(t|s) = \prod_{k=1}^{K+1} \ p(t_k|t_{<k},s,\theta)
    \label{eq:mt_prob_prediction}
\end{equation}

$K$+1 is the length of $t$ with reserved tokens $\langle bos \rangle$ at $k=0$ and $\langle eos \rangle$ at $k$=$|K|$+1 indicating the beginning and end of the prediction.

\subsection{Towards  Controlling  Output Length of MT}
\label{subsec:towards_mt_length_control}
Prior to recent studies on MT output length control~\cite{Lakew19,niehues_machine_2020,lakew2021verbosity}, several attempts have been made for controlling output attributes: text length in summarization~\cite{fan_controllable_2017,takase_positional_2019}, text complexity and formality~\cite{agrawal_controlling_2019,sennrich2016controlling,niu2020controlling}, domain information~\cite{hoang_improved_2018}, and languages in multilingual MT~\cite{johnson2016google}.   
Specific to MT output length control, \cite{Lakew19,niehues_machine_2020} proposed variants of length-encoding motivated by the positional-encoding of self-attention~\cite{vaswani2017attention}. In  closely related works,~\cite{liu2020adapting,karaknta2020subtitle} proposed approaches to respectively control verbosity of speech transcription and speech translation for subtitling. We will focus on approaches that are particularly suited for the \ad~use case.



\subsubsection{Length Token Embedding}
The MT length control approach in~\cite{Lakew19} first proposed to \textit{classify} the bi-text $\mathcal{D}$ 
into three classes, based on the target to source character count length ratio ($LR$) of each sample ($s,t$) pair. 
The class labels are defined on $LR$ thresholds: {\it short} $< 0.95 \geq$ \textit{normal} $\leq 1.05>$ \textit{long}. 
Then model training is performed by prepending the length token $v\in \{short, normal, long\}$, at the beginning of the source sentence. At time of inference, the desired $v$ is prepended on the input sentence.\footnote{Both~\cite{Lakew19} and~\cite{lakew2021verbosity} stated $v=normal$ as the best setting for generating more suitable sentence for AD.} 

In comparison with a baseline length penalty approach to generate shorter translations~\cite{wu2016google}, and the length-encoding variants of~\cite{Lakew19,niehues_machine_2020}, length token has been shown to generate more suitable translations for AD~\cite{lakew2021verbosity}.

\subsubsection{Hypotheses Re-Ranking}
To maximize the suitability of translations for AD,~\cite{saboo_integration_2019} proposed MT hypotheses re-ranking. The approach works by first generating $N$-best hypotheses ($t$) for a source $s$, followed by a re-ranking step based on a scoring function that combines the likelihood of each hypothesis with a so called synchrony score ($S_p$), as follows:

\begin{equation}
  S_d(t,s) = (1-\alpha) \log P(t\mid s) + \alpha S_p(t,s),
    \label{eq:nbest-rescoring}
\end{equation}
Where $S_p$ is computed as~$S_p (t,s)=(1+|len(t)-len(s)|)^{-1}$, $\alpha$ is a parameter tuned on a validation set.

\subsubsection{Length Token + Re-Ranking}
Recently,~\cite{lakew2021verbosity} proposed to combine the length token of~\cite{Lakew19} and a variant of Eq.~\ref{eq:nbest-rescoring} by modifying the synchrony sub-score as, $S_p (t,s) = (1+\frac{len(t)}{len(s)})^{-1}$, which aligns with the overall objective of reducing target to source $LR$ in MT output length control. Fig.~\ref{fig:concise_mt} (\textit{left}), illustrates the approach of \cite{lakew2021verbosity} which reported a higher $\%$ of dubbing suitable translations and human preference of dubbed videos.

\section{Isometric MT with Self-Learning}
\label{sec:concise_mt_selflearning}
In this section, we describe our self-learning based output length control for \mt. To show the impact of our approach, we take \ad~architecture proposed in~\cite{federico_speech--speech_2020}.

\subsection{Self Learning for Isometric MT}
In \mt, the main idea of self-learning is to learn a better model using the predictions of the model itself or an auxiliary 
model output. Self-learning has shown impressive results in unsupervised~\cite{lample2019cross}, zero-shot~\cite{lakew2018improving,zhang2020improving}, and semi-supervised~\cite{sennrich2015improvingMono,edunov2018understanding} neural \mt~modeling. 
Despite the success of self-learning in \mt, existing approaches rely on using external monolingual data and do not evaluate the impact of the pseudo generated data quality before incorporating it into the training stage. 

In this work, the self-learning approach is different in the following aspects, {\it i}) we propose a new synthetic data generation technique based on an output length control criteria, {\it ii}) we then filter and classify the pseudo data based on predefined length classes. In the sections below, we discuss in detail two type of self-learning approaches.

\subsection{Offline Self-Learning}
\label{subsec:offline_sl}
As shown in Fig.~\ref{fig:concise_mt} ({\it center}), the offline self-learning approach considers two independent \mt~models. The first one is a reverse {\it target $\rightarrow$ source} model ({\sc MT$_R$}) trained with the length token approach~\cite{lakew2021verbosity}, optimizing the objective in Eq.~\ref{eq:mt_max_likelihood}. The second model is the desired {\it source $\rightarrow$ target} forward model ({\sc MT$_F$}) which is used to generate isometric translations.

Specifically, for parallel training data $\mathcal{D}$ 
we implement the offline self-learning by first \textit{generating} pseudo source $s'$ from the target $t$, using {\sc MT$_R$}. For inference we prepend $v=long$ on $t$ with the goal of generating more samples with a lower $LR$. 
Ultimately, we aim to address the imbalance of the three length classes in the original bi-text $\mathcal{D}$.

After performing inference for all $t$, we then {\it construct} the pseudo bi-text $\mathcal{D'}=\{(s^{'}_{i}, t_i)$: $ i=1,\ldots,M\}$, 
with length class label for each sample following the LR and thresholds defined in~\cite{lakew2021verbosity}. We then concatenate the original $\mathcal{D}$ and the pseudo $\mathcal{D'}$ bi-text to fine-tune the desired {\sc MT$_F$} model by optimizing Eq.~\ref{eq:mt_max_likelihood}.

\subsection{Online Self-Learning}
\label{subsec:online_sl}
Our second approach primarily avoids learning {\sc MT$_R$} in isolation, as shown in Fig.~\ref{fig:concise_mt} ({\it right}).
We implement a bi-directional model with length classes in each direction ($v\in\{short, normal, long\}$). We prepend a token ($l$) indicating the \textit{source language} on the encoder, and \textit{target language+length} ($lv$) token on the decoder side. For model training, the objective in Eq.~\ref{eq:mt_max_likelihood} is formalized as,

\begin{equation}
\begin{aligned}
     L(\theta) = \argmax{\theta} log\ p(t|s,\theta) + \argmax{\theta} log\ p(s|t,\theta) 
    \label{eq:mt_max_likelihood_online}
\end{aligned}
\end{equation}

During training, we follow a similar three step procedure of the offline self-learning: \textit{length controlled inference, classification of pseudo bi-text into classes, and model training}. 
Differently from the offline, the steps are executed on-the-fly at time of model training and the learning procedure is applied both for the reverse and the forward directions.\footnote{Although, we can evaluate online isometric \mt~both for \textit{source $\leftrightarrow$ target} directions, for a fair comparison with previous work and the offline setting we focus on the forward direction, and leave the rest for future work.} 
We hypothesize modeling a bi-directional isometric \mt~not only able to perform what normally requires two separate MT models, it also positively re-enforce output length control in both directions, with the addition of the pseudo $\mathcal{D'}$ bi-text to balance the length classes in $\mathcal{D}$.

For self-learning approaches, based on how the original ($\mathcal{D}$) and pseudo bi-text ($\mathcal{D'}$) is merged, we have two data configurations, \textit{i}) \textit{Union} = $\mathcal{D} \cup \mathcal{D'}$, and \textit{ii}) \textit{Filter} = $\mathcal{D} \cup \mathcal{D'}$ if LR($t/s'$) $\leq$ 1.05. In other word, for \textit{Filter} we remove the pseudo bi-text data portion labeled  $v=long$.

\section{Experiments}
\label{sec:experiments}
\subsection{Language and Data Processing}
We conduct experiments for English (En) $\rightarrow$ Italian (It), French (Fr), German (De), Spanish (Es) pairs. To avoid the effect of training with low-resource data, we pre-train strong \mt~models on internal data in the magnitude of $10^7$ samples. We then apply our approaches as a fine-tuning step using public data of Ted Talks speeches of $\approx200k$ samples per pair from MuSTC corpus~\cite{MUSTC:2019}. Across all the approaches compared we use the same pre-training and fine-tuning data configuration. Moses scripts 
are used to pre-process the raw data, followed by token segmentation using SentencePiece 
with $32k$ merge operations.\footnote{Moses: \url{https://github.com/moses-smt/mosesdecoder}, SentencePiece:  \url{https://github.com/google/sentencepiece}}

\subsection{Model and Training Configuration}
We use a transformer architecture~\cite{vaswani2017attention} with 6 encoder and decoder layers, self-attention dimension $1024$ and $16$ heads, and feed-forward sublayers of dimension $4096$.
Model is optimized with Adam~\cite{kingma2014adam}, with an initial learning rate of $1\times10^{-7}$. 
Dropout is uniformly set to $0.1$ for all model training. For the online self-learning we share all model parameters between the forward and reverse directions. For model pre-training the best checkpoint is selected based on the lowest loss on the validation set after the training converges. For fine-tuning we train for 5 epochs and take the best checkpoint for evaluation. At inference time we use beam size of $5$, except for the re-ranking approach~\cite{lakew2021verbosity} where beam size $=50$.

\subsection{Model Evaluation}
\label{sec:evaluation}
To evaluate both self-learning approaches, we use the human annotated benchmark from~\cite{virkar2021pa,lakew2021verbosity}. The test set includes $620$ samples per language pair, where references are post-edited to match the source length.

\subsubsection{Automatic Metrics}
We report Sacrebleu scores
~\cite{post-2018-call} on the de-tokenized translated segments and reference pair, target to source length ratio (LR) in character count, and the $\%$ of length compliant ($LC$) translations 
satisfying the $\pm 10\%$ range.
We also report a single score that takes into account both translation quality and length compliance, by simply multiplying the BLEU with LC by which we name \textit{LCB}.

\subsubsection{Human Evaluation}
To align \mt~human evaluation with \ad, we devise an audio based \mt~evaluation scheme. 
We conduct the evaluation by generating audio for a translated segment and its reference using a TTS module. For a fair comparison, a uniform speaking rate is used to generate audio. We then ask annotators to listen to the audio of the translation and the reference and classify them in three categories:

\begin{table}[h!]
     \footnotesize
    \centering
    \tabcolsep=0.18cm
    \begin{tabular}{r|l}
         \textit{Acceptable} ($A$)      & Meaning is similar, fluency is good.  \\
         \textit{Fixable} ($F$)         & Meaning is similar, fluency is poor. \\ 
         \textit{Wrong} ($W$)           & Meaning is different.
    \end{tabular}
\end{table}

To assess overall MT quality the $\%$ of $A$, $F$, and $W$ are computed, whereas to identify length compliant translation we report $LC$ for each rating category. As a final \mt~human evaluation score we compute,

\begin{equation}
HE_{MT} = LC(A) + \frac{1}{2} LC(F)
\label{eq:he_mt}
\end{equation}
$HE_{MT}$ disregards the wrong translations and considers the $\%$ of $A$ and $F$ translations that are isometric. We assume that \textit{acceptable} translations will not require any post-editing, while \textit{fixable} will require minimal post-editing to fix the fluency as such we weigh it with $0.5$.


\section{Result and Discussion}
\label{sec:results}
We compare proposed approaches against a strong \textit{Baseline}, and two MT length control approaches ($v$Tok~\cite{Lakew19} and $v$Tok+Rank~\cite{lakew2021verbosity}). As described in Sec.~\ref{subsec:towards_mt_length_control}, $v$Tok+Rank combines length token ($v$Tok) and hypotheses re-ranking to generate a higher $\%$ of translations suitable for AD~\cite{lakew2021verbosity}. 

\subsection{Automatic Evaluation}
\begin{table}[t!]
    \footnotesize
    \centering
    \tabcolsep=0.21cm
    \begin{tabular}{cccrrrr}
         Pair   & Method        &SL Data            & BLEU      &LR         & LC    & {\bf LCB}   \\ \toprule
                & Baseline      &-                  &46.1      &1.16       &33.1       &15.2 \\ \cline{4-7}
                & $v$Tok        &-                  &\textbf{48.4}      &1.06       &72.7       &35.2   \\
        En-Fr   & $v$Tok+Rank   &-                  &47.7       &1.01       &91.5       &{\bf 43.6}   \\ \cline{2-7}
                & offlineSL     &{\it Union}        &47.9       &1.05       &81.0       &38.8   \\
                &               &{\it Filter}       &47.3     &1.03         &83.4       &39.4   \\ \cline{4-7}
                & onlineSL      &{\it Union}        &48.3     &1.04         &88.9       &\textit{42.9}   \\
                &               &{\it Filter}       &\textbf{48.4}      &1.04        &88.5       &\textit{42.9}   \\ \midrule
                %
                & Baseline      &-                  &37.1       &1.06       &56.5       &20.9   \\ \cline{4-7}
                & $v$Tok        &-                  &37.5       &1.05       &74.7       &28.0   \\
        En-It   & $v$Tok+Rank   &-                  &\textbf{38.5}          &1.00       &90.0       &\textbf{34.7}   \\ \cline{2-7}
                & offlineSL     &{\it Union}        &38.0       &1.04       &80.0       &30.4   \\
                &               &{\it Filter}       &38.1       &1.03       &81.6       &\textit{31.1 }  \\ \cline{4-7}
                & onlineSL      &{\it Union}        &36.2       &1.04       &85.2       &30.8   \\
                &               &{\it Filter}       &36.5       &1.03       &84.7       &30.9   \\ \midrule
                %
                & Baseline      &-                  &33.2       &1.15       &29.8       &9.9   \\ \cline{4-7}
                & $v$Tok        &-                  &34.3       &1.05       &79.2       &27.2   \\
        En-De   & $v$Tok+Rank   &-                  &33.9       &1.00       &93.1       &31.5   \\ \cline{2-7}
                & offlineSL     &{\it Union}        &34.3       &1.01       &86.5       &29.7   \\
                &               &{\it Filter}       &\textbf{34.7}       &1.03       &92.4       &\textbf{32.1}   \\ \cline{4-7}
                & onlineSL      &{\it Union}        &34.0       &1.03       &90.8       &30.9   \\
                &               &{\it Filter}       &34.1       &1.03       &90.5       &30.9   \\ \midrule
                %
                & Baseline      &-                  &\textbf{50.1}       &1.07       &56.0       &28.0   \\ \cline{4-7}
                & $v$Tok        &-                  &49.9       &1.04       &87.6       &43.7   \\
        En-Es   & $v$Tok+Rank   &-                  &49.3       &1.00       &95.5       &47.1   \\ \cline{2-7}
                & offlineSL     &{\it Union}        &50.0       &1.03       &90.8       &45.4   \\
                &               &{\it Filter}       &49.9       &1.03       &93.1       &46.4   \\ \cline{4-7}
                & onlineSL      &{\it Union}        &49.9       &1.03       &95.6       &\textbf{47.7}   \\
                &               &{\it Filter}       &49.8       &1.04       &93.9       &46.7   \\ \bottomrule
    \end{tabular}
    \caption{Results of self-learning isometric MT approaches (\textit{offlineSL}, \textit{onlineSL}) in comparison with previously proposed {\sc mt} output length control mechanisms ($v$Tok~\cite{Lakew19}, and $v$Tok+Rank~\cite{lakew2021verbosity} with N-best hypotheses re-ranking). 
    We measure translation quality (BLEU), length ratio (\textit{LR}), $\%$ of length compliant (\textit{LC}), 
    and overall metric \textit{LCB}. Metric in bold shows best performing approach.}
    \label{tab:main_results}
\end{table}
MT quality and verbosity control are measured using the automatic metrics in Sec.~\ref{sec:evaluation}. 
We observe that length ratio of the Baseline models for En-Fr and En-De is significantly higher than that of En-It and En-Es. This is due to the fact that original MuST-C training data is already well balanced for En-It and En-Es in the ratio of 30/35/35 on an average for short/normal/long classes whereas the same ratio is 15/31/54 for En-Fr and En-De. 
Given that $v$Tok+Rank optimizes over N-best candidate translations, it shows better length control over the Baseline and $v$Tok approach.

Self-learning approaches (\textit{offlineSL} and \textit{onlineSL}), provides consistent improvements over the baseline and $v$Tok in terms of LCB metric for both data configuration (\textit{Union, Filter}). We attribute the performance gains to training models using the additional pseudo parallel data. In fact, performance of self-learning approaches are similar to $v$Tok+Rank for En-De and En-Es which shows how effective controlled self-learning could be, without a hypotheses re-ranking module as in $v$Tok+Rank.
Although, there's no clear winning system in terms of LCB, we prefer onlineSL with Filter simply because it has a higher length compliance (LC) over offlineSL system and Filter has higher LCB on an average when compared to Union approach.


\subsection{Human Evaluation}
Following the evaluation criteria in Sec.~\ref{sec:evaluation}, Table~\ref{tab:mt_human_eval} show results of MT human evaluation. We use 100 randomly selected samples of the test set graded by 40 subjects per language pair and head to head comparison.
We compare our proposed approaches against $v$Tok and 
$v$Tok+Rank. As it turns out \textit{onlineSL} is better than $v$Tok in all languages except En-Fr. This is a very promising result because in one of the language direction (En-Es), we even outperform the $v$Tok+Rank system.

\subsection{Ablation Study}
We also ran an ablation study to check if hypotheses re-ranking complements the self-learning approach. 
We observe that onlineSL+Rank is up to 9\% better than the onlineSL system in terms of relative \textit{LCB} improvements on En-It as shown in Table~\ref{tab:ablation_results}. In fact, onlineSL+Rank beats the state of the art $v$Tok+Rank on three language pairs (except En-It) in terms of LCB. 
Concerning the specific metrics there are only two instances where $v$Tok+Rank outperformed onlineSL+Rank, BLEU for En-It (38.5 vs. 37.0) and LC for En-De (93.1 vs. 92.1). Overall our analysis shows that re-ranking adds on top of the self-learning approach.

\begin{table}[t!]
\centering
\footnotesize
\tabcolsep=0.13cm
\begin{tabular}{lrrr|rr} 
     Pair            &$v$Tok         &offlineSL      &onlineSL   &$v$Tok         &$v$Tok+Rank    \\ \hline
    En-Fr           &\textbf{55.1}	        &45.8           &52.0	    &60.9	        &\textbf{67.9}           \\  
    En-It            &58.0	            &60.9	        &\textbf{66.1}	    &66.6	        &\textbf{73.0}   \\  
    En-De            &67.6	        &66.5	        &\textbf{74.0}	        &65.8	        &\textbf{75.9}       \\   
    En-Es            &69.5	        &67.4	        &\textbf{72.9}	    &66.9	        &\textbf{70.9 }      \\  \hline
\end{tabular}
\caption{Results of head to head MT human evaluation for, $v$Tok Vs. offlineSL Vs. onlineSL, and $v$Tok Vs. $v$Tok+Rank. 
$HE_{MT}$ score is computed using Eq.~\ref{eq:he_mt} that considers acceptable and fixable length compliant translations.}
\label{tab:mt_human_eval}
\end{table}

\begin{table}[t!]
    \footnotesize
    \centering
    \tabcolsep=0.21cm
    \begin{tabular}{ccrrr}
         Pair   & Method                & BLEU  &LC      & {\bf LCB}   \\ \toprule
                & $v$Tok+Rank           &47.7   &91.5     &43.6   \\ 
        En-Fr   & onlineSL              &48.4   &88.5     &42.9  \\ 
                & $+$Rank               &48.4   &93.0     &\textbf{45.0}               \\ \hline
                & $v$Tok+Rank           &38.5   &90.0    &\textbf{34.7}   \\
        En-It   & onlineSL              &36.5   &84.7     &  30.9    \\ 
                & $+$Rank               &37.0   &91.1     &33.7       \\ \hline
                & $v$Tok+Rank           &33.9   &93.1    &31.5    \\ 
        En-De   & onlineSL              &34.1   &90.5    &30.9   \\ 
                & $+$Rank               &34.4   &92.1    &\textbf{31.7}       \\ \hline
                & $v$Tok+Rank           &49.3   &95.5    &47.1     \\ 
        En-Es   & onlineSL              &49.8   &93.9    &46.7  \\ 
                & $+$Rank               &49.3   &98.4    &\textbf{48.5}       \\ \hline
    \end{tabular}
    \caption{Ablation study showing improvement for onlineSL in all language pairs. The results show re-ranking is additive and these approaches are orthogonal.}
    \label{tab:ablation_results}
\end{table}

\section{Conclusion}
In this work, we propose self-learning based approaches to learn an MT model that can generate isometric translations (\textit{i.e., translations matching the source in character count}). 
We evaluate the proposed approach on an automatic dubbing use case, where script translation is expected to match the source length to achieve a synchronicity between source and synthetic target language utterances. Our findings both from automatic and subjective human evaluations show the proposed approach can perform better than a strong model with MT output length control and is on par with the current state of the art that requires generating multiple hypothesis and re-ranking. 

\bibliographystyle{IEEEbib}
\bibliography{paper}

\end{document}